%% file: main.tex
\documentclass[conference]{IEEEtran}
\IEEEoverridecommandlockouts
\usepackage{cite}
\usepackage{amsmath,amssymb,amsfonts}
\usepackage[square,sort,comma,numbers]{natbib}
\usepackage{cite}
\usepackage{booktabs}
\usepackage{algorithmic}
\usepackage{graphicx}
\usepackage{textcomp}
\usepackage{xcolor}
\def\BibTeX{{\rm B\kern-.05em{\sc i\kern-.025em b}\kern-.08em
    T\kern-.1667em\lower.7ex\hbox{E}\kern-.125emX}}
\begin{document}

\title{Explicit Mutual Information Maximization for
Self-Supervised Learning
}


\author{
\IEEEauthorblockN{Lele Chang$^{1}$, Peilin Liu$^{1}$, Qinghai Guo$^{2}$, Fei Wen$^{1}$ \\
\thanks{
}}
\IEEEauthorblockA{
\textit{$^{1}$School of Electronic Information and Electrical Engineering, Shanghai Jiao Tong University, Shanghai, China} \\
\textit{$^{2}$Advanced Computing and Storage Laboratory, Huawei Technologies Co., Ltd., Shenzhen, China}
}
}
\maketitle

\begin{abstract}
Recently, self-supervised learning (SSL) has been extensively studied. Theoretically, mutual information maximization (MIM) is an optimal criterion for SSL, with a strong theoretical foundation in information theory. However, it is difficult to directly apply MIM in SSL since the data distribution is not analytically available in applications. In practice, many existing methods can be viewed as approximate implementations of the MIM criterion. This work shows that, based on the invariance property of MI, explicit MI maximization can be applied to SSL under a generic distribution assumption, i.e., a relaxed condition of the data distribution. We further illustrate this by analyzing the generalized Gaussian distribution. Based on this result, we derive a loss function based on the MIM criterion using only second-order statistics. We implement the new loss for SSL and demonstrate its effectiveness via extensive experiments.
\end{abstract}

\input{contents/intro}
\input{contents/method}
\input{contents/exps/Experiments}

\input{contents/conclusions}
\bibliography{IEEEabrv,DeepLearning.bib}

\onecolumn
\renewcommand{\appendixname}{Supplementary Material}

\begin{appendix}
\input{SM/relatedwork}
\input{SM/ProofTh1}

\input{SM/Ablations}
\input{SM/exp_details}
\end{appendix}



\end{document}

%% file: contents/intro.tex
\section{Introduction}\label{sec:introduction}
Self-supervised learning (SSL) is a powerful technique 
aims to learn task-agnostic representations without relying on annotated training data.
SSL methods commonly utilize pretext tasks to learn 
representation-extracting models,
which can be generally categorized into 
generative-based and contrastive-based ones \cite{kimHypeBoyGenerativeSelfSupervised2024,jaiswalSurveyContrastiveSelfSupervised2021}. 
In generative-based methods, the pretext tasks are  designed to predict or reconstruct the input data \cite{liuSelfSupervisedLearningTime2024,heMaskedAutoencodersAre2021}. 
In comparison, contrastive-based methods construct pretext tasks in the embedding space, and learn a representation-extracting model by discriminating between positive and negative pairs in the embedding space.

Recently, Siamese networks \cite{bromley1993signature} based 
SSL methods have shown great promise,
which utilize weight-sharing networks to 
process distorted versions of the samples 
\cite{chenSimpleFrameworkContrastive2020,zbontarBarlowTwinsSelfSupervised2021,bardesVICRegVarianceInvarianceCovarianceRegularization2022,caronUnsupervisedLearningVisual2020,heMomentumContrastUnsupervised2020}. 
An issue of Siamese networks is 
that the two branches may collapse to constant representation.
To prevent this problem, contrastive based methods construct positive and negative pairs and contrast between them using  contrastive losses, such as InfoNCE loss \cite{oordRepresentationLearningContrastive2019}, 
NT-Xent loss \cite{chenSimpleFrameworkContrastive2020}, Circle loss \cite{sunCircleLossUnified2020}, 
and Triplet loss \cite{hofferDeepMetricLearning2018}.
Besides, 
online clustering, asymmetric structure and momentum encoder have been shown to be effective for preventing collapsing  \cite{caronUnsupervisedLearningVisual2020,grillBootstrapYourOwn2020,chenExploringSimpleSiamese2020,heMomentumContrastUnsupervised2020}.
Furthermore, 
redundancy reduction of the embeddings and variance regularization 
can naturally avoid collapsing \cite{zbontarBarlowTwinsSelfSupervised2021,bardesVICRegVarianceInvarianceCovarianceRegularization2022}.
Generally, the primary objective of these methods is to maximize the similarity between the two-branch embeddings of the Siamese networks while avoiding the collapsing problem.  

Mutual information (MI) is an information-theoretic measure that can capture non-linear statistical dependencies between random variables and thereby serve as a measure of true dependence \cite{kinney2014equitability}.
Theoretically, under the framework of Siamese networks based SSL, MI is an optimal metric to measure the dependence between the embeddings of the Siamese networks. However, directly constructing an objective based on MI is challenging as it relies on the analytic expression of the data distribution. Consequently, many existing methods can be seen as approximate implementations of the MI maximization criterion. For example, the InfoNCE loss based methods \cite{chenSimpleFrameworkContrastive2020} 
employ a noise contrastive estimation (NCE) of MI \cite{gutmann2010noise}, which is shown to be a lower bound of MI \cite{poole2019variational}. Besides, it has been shown in \cite{shwartz2023information} that the VICReg method  \cite{bardesVICRegVarianceInvarianceCovarianceRegularization2022} implements an approximation of the MI maximization criterion. 
In \cite{liuSelfSupervisedLearningMaximum2022}, 
the minimal coding length in lossy coding is used to construct a maximum entropy coding objective for SSL.
Meanwhile, MI maximization is considered in \cite{ozsoy2022self} for
SSL using a log-determinant approximation of the MI,
which is simplified to derive a Euclidean distance-based objective.

In this work, based on the invariance property of MI, we show that the explicit MI maximization objective based on second-order statistics can be applied to SSL under a generic condition of the data distribution. This objective involves intractable determinant calculation of high-dimensional matrices, we propose reformulation and optimization strategies to make it adequate for stable and efficient end-to-end training.
The main contributions are  as follows:
\begin{itemize}
\item 
We provide a new perspective for SSL objective design from the 
invariance property of MI established in information theory,
which shows that explicit MI optimization based on second-order statistics can be applied 
to SSL under a generic condition of the data distribution. 
\item
We implement the second-order statistics based MI objective for SSL, for which optimization strategies have been
proposed to make it adequate for end-to-end self-supervised learning on practical tasks.
\item
We demonstrate the effectiveness of the proposed method 
by evaluation on CIFAR-10/100 and ImageNet-100/1K 
in comparison with state-of-the-art methods.

\end{itemize} 

%% file: contents/method.tex
\section{Method} \label{sec:method}
\begin{figure}
    \centering
    \includegraphics[width=1\linewidth]{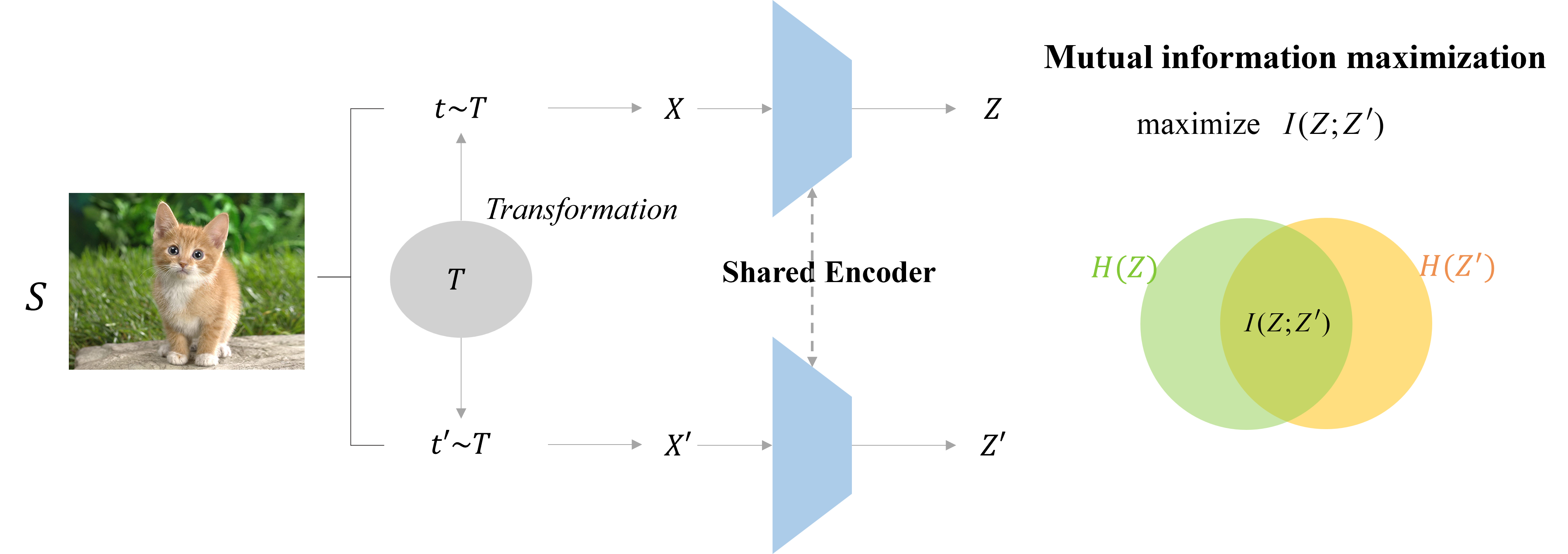}
    \caption{
The MMI objective explicitly measures the MI $I(Z;Z^{\prime})$ between the  embeddings $Z$ and $Z^{\prime}$ generated by two identical networks $f_{\omega}(\cdot)$ that are fed transformed versions of sample $S$. Maximizing  $I(Z;Z^{\prime})$ not only maximizes the dependency between the embeddings $Z$ and $Z^{\prime}$ by minimizing their joint entropy $H(Z,Z^{\prime})$, but also maximizes their marginal entropy $H(Z)$ and $H(Z^{\prime})$, respectively, which naturally avoids trivial constant solutions.}
    \label{fig:Siam_pipline}
    \vspace{-4mm}
\end{figure}

We consider a Siamese network architecture that processes two augmented versions of the same input through parallel identical networks \cite{chiccoSiameseNeuralNetworks2021a}, as illustrated in Fig. \ref{fig:Siam_pipline}. 
Denote the two embeddings by $Z  =\left[Z_{1}, Z_{2}, \cdots, Z_{d}\right]^{T} \in \mathbb{R}^{d}$ and $Z^{\prime}  =\left[Z_{1}^{\prime}, Z_{2}^{\prime}, \cdots, Z_{d}^{\prime}\right]^{T} \in \mathbb{R}^{d}$, respectively, which are generated from two distorted views $X$ and $X^{\prime}$ of the same image $S$ by the network $f_{\omega}(\cdot)$ with trainable parameters $\omega$.

\subsection{The Maximum Mutual Information Criterion}

As shown in Fig. \ref{fig:Siam_pipline}, we consider 
a maximum mutual information (MMI) criterion as  
\begin{equation*}\label{MMI-formulation}
\text { maximize } I\left(Z ; Z^{\prime}\right). \tag{1}
\end{equation*}
The MI can be expressed as $I\left(Z ; Z^{\prime}\right)=H(Z)+H(Z^{\prime})-H(Z, Z^{\prime})$,
where $H(Z, Z^{\prime})$ is the joint entropy of $(Z, Z^{\prime})$, $H(Z)$ and $H(Z^{\prime})$ are the marginal entropy of $Z$ and $Z^{\prime}$, respectively. The MMI criterion not only maximizes the dependency between the embeddings $Z$ and $Z^{\prime}$ by minimizing their joint entropy, 
but also maximizes the marginal entropy of $Z$ and $Z^{\prime}$, respectively, which promotes decorrelating the features in the embeddings $Z$ and $Z^{\prime}$ and naturally avoids the collapse problem of converging to trivial constant solutions. 



However, MI is difficult to explicitly compute except for certain simple specific distributions. Accordingly, Gaussian distribution assumption is used to derive MI based objectives for SSL \cite{ozsoy2022self,shwartz2023information}. The next result shows that $I\left(Z ; Z^{\prime}\right)$ can be explicitly computed based only on second-order statistics under a more generic distribution condition. In the following, we denote the covariance matrices of $Z, Z^{\prime}$ and $\tilde{Z}:=\left[Z^{T}, Z^{\prime T}\right]^{T}$ by $C_{Z Z}$, $C_{Z^{\prime} Z^{\prime}}$ and $C_{\tilde{Z} \tilde{Z}}$, respectively.

\textit{\textbf{Theorem 1}}: If there exist two homeomorphisms maps $F: \mathbb{R}^{d} \rightarrow \mathbb{R}^{d}$ and $G: \mathbb{R}^{d} \rightarrow \mathbb{R}^{d}$ (i.e. $F$ and $G$ are smooth and uniquely invertible maps) such that $Y=F(Z)$ and $Y^{\prime}=G\left(Z^{\prime}\right)$ are Gaussian distributed, 
let $\tilde Y = {[{Y^T},{Y'^T}]^T}$ and denote the covariance
matrices of $Y$, $Y^{\prime}$ and $\tilde Y$ by $C_{Y Y}$, $C_{Y^{\prime} Y^{\prime}}$ and $C_{\tilde{Y} \tilde{Y}}$, respectively,
then
\begin{equation*}\label{MI-theorem1}
I\left(Z ; Z^{\prime}\right)=I\left(Y ; Y^{\prime}\right)=\frac{1}{2} \log \frac{\operatorname{det}\left(C_{Y Y}\right) \operatorname{det}\left(C_{Y^{\prime} Y^{\prime}}\right)}{\operatorname{det}\left(C_{\tilde{Y} \tilde{Y}}\right)}. \tag{2}
\end{equation*}

The proof is given in the supplemental material (SM) \cite{chang2024}.
This result is derived from the invariance property of MI \cite{kraskov2004estimating} and the MI of multivariate Gaussian distribution.  
It relaxes the condition of data distribution required for explicit MI expression. It implies that we can compute the MI only based on second-order statistics even if the distributions of $Z$ and $Z^{\prime}$ are not Gaussian. To verify this result, we investigate the MI under a class of distributions, namely generalized Gaussian distribution (GGD). Generally, there does not appear to exist an agree on multivariate GGD, we use a definition as a particular case of the Kotz-type distribution \cite{fang2018symmetric, verdoolaege2012geometry} 
\begin{equation*}
\begin{aligned}
 & \mathcal{G N}\left(X ; \mu_{X}, \Sigma_{X X}, \beta\right):= \\ 
& \frac{\Phi(\beta, n)}{\left[\operatorname{det}\left(\Sigma_{X X}\right)\right]^{1 / 2}} \exp \left(-\frac{1}{2}\left[\left(X-\mu_{X}\right)^{T} \Sigma_{X X}^{-1}\left(X-\mu_{X}\right)\right]^{\beta}\right), 
\end{aligned}
\tag{3}
\end{equation*}
%
where $\Gamma$ is the gamma function, $\Phi(\beta, n)=\beta \Gamma(n / 2) /\left[2^{n /(2 \beta)} \pi^{n / 2} \Gamma(n /(2 \beta)]\right.$,  $\mu_{X}$ and $\Sigma_{X X}$ are the mean and dispersion matrix of $X$, respectively, $\beta>0$ is a shape parameter. GGD has a flexible parametric form, which can adapt to a large family of distributions by choosing the shape parameter $\beta$, from super-Gaussian when $\beta<1$ to sub-Gaussian when $\beta>1$, including the Gamma, Laplacian and Gaussian distributions as special cases. 

\vspace{2mm}
\textit{\textbf{Theorem 2}}: 
Suppose that $Z$ and $Z^{\prime}$ follow multivariate GGD, $Z \sim \mathcal{G N}\left(\mu_{Z}, \Sigma_{Z Z}, \beta\right), Z^{\prime} \sim \mathcal{G N}\left(\mu_{Z^{\prime}}, \Sigma_{Z^{\prime} Z^{\prime}}, \beta\right)$, where $\Sigma_{ZZ}$ and $\Sigma_{Z^{\prime} Z^{\prime}}$ are dispersion matrices. Let $\tilde{Z}=\left[Z^{T}, Z^{\prime T}\right]^{T} \in \mathbb{R}^{2 d}$ and denote the dispersion matrix of $\tilde{Z}$ by $\Sigma_{\tilde{Z} \tilde{Z}}$. Then, for any $\beta>0$, the MI between $Z$ and $Z^{\prime}$ is given by
\begin{equation*}
\begin{aligned}
    I\left(Z ; Z^{\prime}\right) &=\frac{1}{2} \log \frac{\operatorname{det}\left(\Sigma_{ZZ}\right) \operatorname{det}\left(\Sigma_{Z^{\prime} Z^{\prime}}\right)}{\operatorname{det}\left(\Sigma_{\tilde{Z} \tilde{Z}}\right)} \\ &=\frac{1}{2} \log \frac{\operatorname{det}\left(C_{ZZ}\right) \operatorname{det}\left(C_{Z^{\prime} {Z}^{\prime}}\right)}{\operatorname{det}\left(C_{\tilde{Z} \tilde{Z}}\right)}. 
\end{aligned} \tag{4}
\end{equation*}
The proof is given in the SM \cite{chang2024}. 
The above result implies that MI can be explicitly computed based only on second-order statistics for any distribution that can be marginally reparameterized as multivariate norm distribution under a homeomorphism condition.

Based on the above results, we propose to use the second-order statistics based explicit MI expression for SSL, with which the MMI criterion (\ref{MMI-formulation}) can be formulated as
\begin{equation*}\label{SSL_MMI_formulation}
\text { minimize } \log \operatorname{det}\left(C_{\tilde{Z} \tilde{Z}}\right)-\log \operatorname{det}\left(C_{ZZ}\right)-\log \operatorname{det}\left(C_{Z^{\prime} Z^{\prime}}\right). \tag{5}
\end{equation*}
Minimizing formulation (\ref{SSL_MMI_formulation}) is equivalent to maximizing $\operatorname{det}\left(C_{Z Z}\right)$ and $\operatorname{det}\left(C_{Z^{\prime} Z^{\prime}}\right)$ while minimizing $\operatorname{det}\left(C_{\tilde{Z} \tilde{Z}}\right)$. Maximizing $\operatorname{det}\left(C_{Z Z}\right)$ (resp. $\operatorname{det}\left(C_{Z^{\prime} Z^{\prime}}\right)$ ) would promote the features in $Z$ (resp. $Z^{\prime}$ ) to be less correlated, therefore decorrelates the features in the output embedding. This decorrelation reduces the redundant information of the output embedding and promotes extracting information from the sample as much as possible, which also helps avoid the collapse problem of converging to a trivial constant solution.

\subsection{Implementation of the MMI Criterion for SSL}
It is difficult to directly optimize the MMI formulation (\ref{SSL_MMI_formulation}) as it involves computing the determinant of high-dimensional covariance matrices, e.g., $C_{ZZ} \in \mathbb{R}^{d \times d}, C_{Z^{\prime} Z^{\prime}} \in \mathbb{R}^{d \times d}$, and $C_{\tilde{Z}\tilde{Z} } \in \mathbb{R}^{2 d \times 2 d}$. Typically, $m<d$ in the self-supervised learning setting, e.g., $d=4096, m \in\{512,1024,2048\}$. The MMI formulation can be reformulated to reduce the computational complexity. Specifically, since the two transformation operators $X=t\left(S\right)$ and $X^{\prime}=t^{\prime}\left(S\right)$ are sampled from the same distribution of augmentations, i.e. $t \sim T$ and $t^{\prime} \sim T, Z=f_{\omega}(X)$ and $Z^{\prime}=f_{\omega}\left(X^{\prime}\right)$ are the representations extracted by the same network $f_{\omega}$, and without loss of generality assuming that $Z$ and $Z^{\prime}$ having zero-mean, it follows that
$$
\begin{aligned}
& C_{Z Z}=E\left\{Z Z^{T}\right\}=C_{Z^{\prime} Z^{\prime}}=E\left\{Z^{\prime} Z^{\prime T}\right\}, \\
& C_{Z Z^{\prime}}=E\left\{Z Z^{\prime T}\right\}=C_{Z^{\prime} Z}=E\left\{Z^{\prime} Z^{T}\right\}.
\end{aligned}
$$
Then, using $\operatorname{det}\left(\left[\begin{array}{ll}\!A & B \!\\ \!B & A\!\end{array}\right]\right)=\operatorname{det}(A+B) \operatorname{det}(A-B)$, we have
$$
\begin{aligned}
    \operatorname{det}\left(C_{\tilde{Z}\tilde{Z}}\right) 
&=\operatorname{det}\left(\left[
\begin{array}{cc}
\!C_{ZZ} & C_{ZZ^{\prime}} \!\\
\!C_{Z^{\prime}Z} & C_{Z^{\prime} Z^{\prime}}\!
\end{array}\right]\right) \\
&=\operatorname{det}\left(C_{ZZ}+C_{ZZ^{\prime}}\right) \operatorname{det}\left(C_{ZZ}-C_{ZZ^{\prime}}\right).
\end{aligned}
$$
Therefore, the objective of the MMI formulation (\ref{SSL_MMI_formulation}) can be reformulated as
\begin{equation*}\label{eq6}
\begin{aligned}
    \mathcal{L} &=\log \operatorname{det}\left(C_{ZZ}+C_{ZZ^{\prime}}\right)+\log \operatorname{det}\left(C_{ZZ}-C_{ZZ^{\prime}}\right) \\ &-\log \operatorname{det}\left(C_{ZZ}\right) 
-\log \operatorname{det}\left(C_{Z^{\prime}Z^{\prime}}\right). 
\end{aligned}\tag{6}
\end{equation*}

In the following without loss of generality we assume that $Z$ and $Z^{\prime}$ are normalized such that $\operatorname{diag}\left\{C_{ZZ}\right\}=\operatorname{diag}\left\{C_{Z^{\prime} Z^{\prime}}\right\}=I$. Then, it is easy to see that, the desired optimal solution of minimizing (\ref{eq6}) is $C_{ZZ}= \pm C_{ZZ^{\prime}}$ and $C_{ZZ}=C_{Z^{\prime}Z^{\prime}}=I$. That is $Z= \pm Z^{\prime}$, and, at meantime, $Z$ and $Z^{\prime}$ have the maximum entropy given by $C_{ZZ}=C_{Z^{\prime}Z^{\prime}}=I$. However, from the above analysis, since $X=t\left(S\right)$ and $X^{\prime}=t^{\prime}\left(S\right)$ with $t(\cdot)$ and $t^{\prime}(\cdot)$ following the same distribution, $Z=f_{\omega}(X)$ and $Z^{\prime}=f_{\omega}\left(X^{\prime}\right)$ are the outputs of the same network $f_{\omega}$, the solution $C_{ZZ}=-C_{ZZ^{\prime}}$ given by $Z=-Z^{\prime}$ is invalid. Hence, we can drop the first term in (\ref{eq6}) to obtain
\begin{equation*}\label{SSL_MMI_formulation2}
\mathcal{L}=\log \operatorname{det}\left(C_{ZZ}-C_{ZZ^{\prime}}\right)-\log \operatorname{det}\left(C_{ZZ}\right)-\log \operatorname{det}\left(C_{Z^{\prime} Z^{\prime}}\right). \tag{7}
\end{equation*}

It is problematic to directly optimize the objective (\ref{SSL_MMI_formulation2}). First, the MI is formulated for multivariate continuous variables, which would be infinite when any two components are linearly correlated, e.g., $I\left(Z ; Z^{\prime}\right)=\infty$ if $Z_{i}=\phi Z_{i}^{\prime}$ for any $i \in\{1, \cdots, d\}$ and any $\phi \neq 0$. This means $\mathcal{L}=-\infty$ when any pair $\left(Z_{i}, Z_{i}^{\prime}\right)$ is linearly correlated, regardless the correlation between other feature pairs $\left\{\left(Z_{j}, Z_{j}^{\prime}\right)\right\}_{j \neq i}$. Second, the high-dimensional determinant is intractable to compute, and direct minimization of (\ref{SSL_MMI_formulation2}) has a degeneration problem when $m<d$ since the covariance matrices are rank-deficient in this case. To achieve stable optimization for end-to-end training, we consider further reformulation and approximation of (\ref{SSL_MMI_formulation2}) as follows.
Specifically, in implementation the covariance matrices are estimated from $m$ samples, and an empirical loss of (\ref{SSL_MMI_formulation2}) is used as
\begin{equation*}\label{SSL_MMI_formulation4}
\mathcal{L}=\log \operatorname{det}(\hat{C}_{Z Z}-\hat{C}_{Z Z^{\prime}})-\log \operatorname{det}(\hat{C}_{ZZ})-\log \operatorname{det}(\hat{C}_{Z^{\prime} Z^{\prime}}). \tag{8}
\end{equation*}
With $m<d$, the matrices $\hat{C}_{ZZ}, \hat{C}_{ZZ^{\prime}}, \hat{C}_{Z^{\prime} Z^{\prime}}$ are rank-deficient and their determinants are constantly zero, which makes (\ref{SSL_MMI_formulation4}) an inadequate objective for optimizing the network. In this case, recalling that the determinant of a matrix is equivalent to the product of its eigenvalues, we turn to optimize the product of its nonzero eigenvalues. 
Under the assumption that $Z$ and $Z^{\prime}$ have zero-mean, and let $\Bar{Z} \in \mathbb{R}^{d \times m}$ and $\Bar{Z}^{\prime} \in \mathbb{R}^{d \times m}$ denote the matrices containing the embeddings of $m$ samples,
this can be implemented by replacing $\hat{C}_{Z Z}=\frac{1}{d} \Bar{Z} \Bar{Z}^{T}, \hat{C}_{Z Z^{\prime}}=\frac{1}{d} \Bar{Z} \Bar{Z}^{\prime T}, \hat{C}_{Z^{\prime} Z^{\prime}}=\frac{1}{d} \Bar{Z}^{\prime} \Bar{Z}^{\prime T}$ by $\breve{C}_{Z Z}=\frac{1}{m} \Bar{Z}^{T} \Bar{Z}, \breve{C}_{Z Z^{\prime}}=\frac{1}{m} \Bar{Z}^{T} \Bar{Z}^{\prime}$, and $\breve{C}_{Z^{\prime} Z^{\prime}}=\frac{1}{m} \Bar{Z}^{\prime T} \Bar{Z}^{\prime}$, respectively, to get
\begin{equation*}\label{SSL_MMI_formulation3}
\mathcal{L}\!=\!\log \operatorname{det}\left(\Bar{Z}^{T} \Bar{Z}-\Bar{Z}^{T} \Bar{Z}^{\prime}\right)-\log \operatorname{det}\left(\Bar{Z}^{T} \Bar{Z}\right)-\log \operatorname{det}\left(\Bar{Z}^{\prime T} \Bar{Z}^{\prime}\right). \tag{9}
\end{equation*}

Besides, since the determinant of a high dimensional matrix is expensive to compute and unstable, we consider a Tylor expansion of the log-determinant $\log \operatorname{det}(M)$ for a matrix $M \in \mathbb{R}^{n \times n}$ as
\begin{align*}\label{tyler_expan}
 & \log \operatorname{det}(M)=\sum_{i=1}^{n} \log \lambda_{i}(M) \\
& =\sum_{i=1}^{n} \sum_{k=1}^{\infty}(-1)^{k+1} \frac{\left(\lambda_{i}(M)-1\right)^{k}}{k} \\
& =\sum_{k=1}^{\infty}(-1)^{k+1} \frac{\operatorname{tr}\left((M-I)^{k}\right)} {k} 
\operatorname{tr}\left(\sum_{k=1}^{\infty}(-1)^{k+1} \frac{(M-I)^{k}}{k}\right),\tag{10}
\end{align*}
where we used $\sum_{i=1}^{n} \left(\lambda_{i}(M)-1\right)^{k}=\operatorname{tr}\left((M-I)^{k}\right)$, and $\lambda_{i}(M)$ denotes the $i$-th eigenvalue of $M$. The series (\ref{tyler_expan}) only converges under the condition of $\|M-I\|<1$, or equivalently $0<\lambda_{i}(M)<2$ for any $i$. We seek to use a low-order approximation of (\ref{tyler_expan}) to facilitate the optimization. From (\ref{tyler_expan}), since a lower-order approximation is more accurate when the eigenvalues $\left\{\lambda_{i}(M)\right\}_{i=1, \ldots, n}$ are closer to 1, we rescale the matrix as
\begin{equation*}\label{rescaling}
\tilde{M}=\frac{M-\mu_{\lambda} I}{\alpha}+I, \tag{11}
\end{equation*}
where $\mu_{\lambda}=\frac{\lambda_{\max }(M)+\lambda_{\min }(M)}{2}$ and $\alpha=\beta\left(\mu_{\lambda}-\lambda_{\text {min }}(M)\right)$, such that $\|\tilde{M}-I\| \leq \frac{1}{\beta}<1$ for some $\beta>1.
~\lambda_{\text {min }}(\cdot)$ and $\lambda_{\text {max }}(\cdot)$ denote the minimum and maximum eigenvalues of a matrix, respectively. In implementing the loss (\ref{SSL_MMI_formulation3}), the three log-determinant terms are expanded as (\ref{tyler_expan}), and only a $p$-th order approximation is kept, e.g., $p=4$ in the experiments of this work. A fourth-order approximation of the log function in (\ref{tyler_expan}) is sufficiently accurate around the value of 1.

Notice that using the diagonal-loading like rescaling in (\ref{rescaling}) is equivalent to using a noise injection regularization of the MI calculation. Specifically, it is equivalent to replacing $I\left(Z ; Z^{\prime}\right)$ by $I\left(Z +\varepsilon; Z^{\prime}+\varepsilon ^\prime\right)$, where $\varepsilon$ and $\varepsilon^\prime$ are additive Gaussian noise.
With this regularization the MI becomes finite.
Otherwise, the MI would be infinite when any two components are linearly correlated, e.g., $I\left(Z ; Z^{\prime}\right)=\infty$ if $Z_{i}=\phi Z_{i}^{\prime}$ for any $i \in\{1, \cdots, d\}$ and any $\phi \neq 0$,
and correspondingly with the first term of the objective (\ref{SSL_MMI_formulation}) becoming minus infinite, which makes the objective in (\ref{SSL_MMI_formulation}) difficult to optimize directly.



%% file: contents/exps/Experiments.tex
\section{Experiments} \label{sec:experiments}
We follow the standard training protocol in Solo-learn benchmark \cite{costaSololearnLibrarySelfsupervised2022}. ResNet18 is used as the backbone for CIFAR-10/100 and ImageNet-100, whilst ResNet50 is used for ImageNet-1K. The results on CIFAR-10/100 and ImageNet-100 in Table \ref{tab:CIFAR_linear} are compared with a batch size of 256.  
The experiments are conducted on a computing platform with three A100 GPUs, each with 40 GB of memory. On ImageNet-1K, it supports a maximum batch size of 1020 when using 16-bit precision (FP16). Therefore, on ImageNet-1K, our method is run with a precision of FP16 and a batch size of 1020. Meanwhile, to enhance stability and expedite the training process, we use gradient accumulation across four batches ($1020\times4$) on ImageNet-1K.
Detailed experiment settings are provided in the SM \cite{chang2024}. 

\begin{table}
\caption{\textbf{Linear probing results on CIFAR-10/100 and ImageNet-100.} 
Top-1 accuracy is reported with ResNet18.
}
  \label{tab:CIFAR_linear}
  \centering
  \begin{tabular}{lccc}
    \toprule
    Method & CIFAR-10 & CIFAR-100 & ImageNet-100 \\
    \midrule
    Barlow Twins \cite{zbontarBarlowTwinsSelfSupervised2021}  & 92.1 & \textbf{70.9} & 80.4 \\
BYOL\cite{grillBootstrapYourOwn2020} & 92.6 & 70.5 & 80.2 \\ 
    DeepCluster V2 \cite{caronUnsupervisedLearningVisual2020} & 88.9 & 63.6 & 75.4 \\
    DINO \cite{caronEmergingPropertiesSelfSupervised2021} & 89.5 & 66.8 & 74.8 \\
    MoCo V2+ \cite{chenImprovedBaselinesMomentum2020} & 92.9 & 69.9 & 78.2 \\
    MoCo V3 \cite{chenEmpiricalStudyTraining2021} & \textbf{93.1} & 68.8 & 80.4 \\
    NNCLR \cite{dwibediLittleHelpMy2021} & 91.9 & 69.6 & 79.8 \\
    ReSSL \cite{zhengReSSLRelationalSelfSupervised2021} & 90.6 & 65.9 & 76.9 \\
    SimCLR \cite{chenSimpleFrameworkContrastive2020} & 90.7 & 65.8 & 77.6 \\
    Simsiam \cite{chenExploringSimpleSiamese2020} & 90.5 & 66.0 & 74.5 \\
    SwAV \cite{jingSelfsupervisedVisualFeature2019} & 89.2 & 64.9 & 74.0 \\
VICReg \cite{bardesVICRegVarianceInvarianceCovarianceRegularization2022} & 92.1 & 68.5 & 79.2 \\
    W-MSE \cite{ermolovWhiteningSelfSupervisedRepresentation2021} & 88.7 & 61.3 & 67.6 \\  
     CorInfoMax \cite{ozsoy2022self} & \textbf{93.2} & \textbf{71.6} & \textbf{80.5} \\
    Ours & \textbf{93.1} & 70.5 & \textbf{81.1} \\
    \bottomrule
\end{tabular}
\end{table}

\begin{table}
 \caption {\textbf{Linear probing results on ImageNet-1K.}  
  Top-1 accuracy is reported at 100, 400, and 800 epochs with ResNet50.}
\label{tab:ImageNet_linear}
  \centering
  \label{tab:linear_probing}
  \begin{tabular}{lccc}
    \toprule
    Method & 100 eps & 400 eps& 800 eps\\
    \midrule
    SimCLR \cite{chenSimpleFrameworkContrastive2020}  & 66.5 & 69.8 & 70.4 \\
    MoCo v2 \cite{chenImprovedBaselinesMomentum2020}& 67.4 & 71.0 & 72.2 \\
    BYOL \cite{grillBootstrapYourOwn2020} & 66.5 & \textbf{73.2} & \textbf{74.3} \\ 
    SwAV \cite{jingSelfsupervisedVisualFeature2019}  & 66.5 & 70.7 & 71.8 \\
    SimSiam \cite{chenExploringSimpleSiamese2020}  & \textbf{68.1} & 70.8 & 71.3 \\
    Barlow Twins \cite{zbontarBarlowTwinsSelfSupervised2021}& 67.3 & 71.8 & 73.0 \\
    Ours & \textbf{70.6} & \textbf{72.6} &\textbf{73.1} \\
    \bottomrule
  \end{tabular}
\end{table}

\paragraph{LINEAR PROBING}Linear probing is a standard evaluation protocol in SSL,
in which a linear classifier is trained on top of frozen 
representations of a pretrained model.
Table \ref{tab:CIFAR_linear} shows the linear probing results on CIFAR-10/100 and ImageNet-100.  
Following strictly the protocol of the Solo-learn benchmark \cite{costaSololearnLibrarySelfsupervised2022}, 
ResNet18 is used with the last fully connected layer replaced with a three-layer MLP interleaved with Batch Normalization (BN) and ReLU as projector. We keep the projector's hidden dimension and output dimension as 2048. The modified backbone and projector act as  encoder. 
For CIFAR, we adapt the architecture by removing the first max-pooling layer and modifying the first convolutional layer to fit the $32 \times 32$ input size, in line with \cite{costaSololearnLibrarySelfsupervised2022,chenSimpleFrameworkContrastive2020}. The results for other methods in Table \ref{tab:CIFAR_linear} are sourced from the Solo-learn benchmark \cite{costaSololearnLibrarySelfsupervised2022}. It can be seen that our method achieves competitive performance on all three datasets. Notably, on ImageNet-100, our method achieves the highest accuracy of 81.1\%. 
Table \ref{tab:linear_probing} compares the top-1 accuracy of the methods on ImageNet-1K with different pretraining epochs. 
On ImageNet-1K, the results of our method are reported with a momentum encoder \cite{grillBootstrapYourOwn2020,chenExploringSimpleSiamese2020}. As shown in Table \ref{tab:abl_bs}, our method achieves better performance with a momentum encoder.
From Table \ref{tab:linear_probing}, our method 
achieves competitive performance on ImageNet-1K,  which is especially noteworthy under small training epochs. 



\paragraph{ABLATION ON BATCH SIZE}
We evaluate the robustness of our method against batch size, 
following the same setting as Table \ref{tab:CIFAR_linear}. 
We consider two variants of our method, one is standard and the
other uses a momentum encoder \cite{grillBootstrapYourOwn2020,chenExploringSimpleSiamese2020}.
As shown in Table \ref{tab:abl_bs}, both our method and Barlow Twins reach a saturation point as batch size increases, which is consistent with the results in \cite{zbontarBarlowTwinsSelfSupervised2021}. Our method has better robustness across varying batch sizes, with a maximum accuracy variation of 1.6\% on CIFAR-100 and 2.0\% on ImageNet-100, respectively, while Barlow Twins shows 5.6\% and 2.9\% accuracy variation on the two datasets. For our method, using a momentum encoder helps to enhance performance across different batch sizes. More ablation studies can refer to the SM \cite{chang2024}.
\begin{table}[!t]
  \caption{\textbf{Robustness to batch size.} 
  We use the same setup as in Table \ref{tab:CIFAR_linear} but vary the batch size. ``Ours-M" denotes our method with a momentum encoder. 
  }
  \label{tab:abl_bs}
  \centering
  \footnotesize
  \begin{tabular}{p{0.4cm} p{0.3cm} p{0.9cm} p{1.6cm} p{0.3cm} p{0.9cm} p{1.6cm}}
    \toprule
    Batch size & \multicolumn{3}{c}{CIFAR-100} & \multicolumn{3}{c}{ImageNet-100} \\
    \cmidrule(r){2-4} \cmidrule(r){5-7}
     & Ours & Ours-M& Barlow Twins& Ours & Ours-M& Barlow Twins\\
    \midrule
    1024 & 68.9 & 70.4 & 65.9 & 79.8 & 81.4 & 77.5 \\
    512 & 70.0 & 70.5 & 68.7 & 79.5 & 81.1& 79.6 \\
    256 & 70.5 & 70.4 & 70.9 & 81.1 & 81.7 & 80.4 \\
    128 & 70.2 & 70.2 & 71.5 & 80.7 & 81.8 & 80.2 \\
    64 & 70.1 & 71.1 & 70.2 & 79.1 & 81.2 & 78.4 \\
    \bottomrule
  \end{tabular}
    \vspace{-4mm}
\end{table}

%% file: contents/conclusions.tex
\section{Conclusions} \label{sec:conclusion}
This work seeks to construct an objective for SSL from the 
information theoretic view. Based on the invariance property of 
mutual information, we showed that the maximum MI criterion can be applied to 
SSL for a relaxed condition on the data distribution. We illustrated
this through analyzing the explicit MI of generalized Gaussian distribution.
Then, the derived second-order statistics based formulation of MI
is employed for SSL, and  optimization strategies have been proposed to efficiently implement it 
for end-to-end training.
Experiments on the CIFAR and ImageNet benchmarks  
demonstrated the effectiveness of the derived MI based loss.


%% file: SM/relatedwork.tex
\subsection{Related Work}
\label{sec:relatedwork}
While there exists a large number of SSL methods developed in
the last years, here we mainly review recent contrastive-based methods closely related to ours. Besides, while our method is based on MI optimization, we particularly review the works related to MI.
\subsubsection{Siamese Networks Based Self-Supervised Representation Learning}
Siamese networks \cite{bromley1993signature} have become 
a prevalent structure in recent SSL models and achieved 
great performance \cite{chenSimpleFrameworkContrastive2020,heMomentumContrastUnsupervised2020,grillBootstrapYourOwn2020,zbontarBarlowTwinsSelfSupervised2021,caronUnsupervisedLearningVisual2020}. 
Siamese structure-based methods utilize two weight-sharing network 
encoders to process distinct views of the same input image,
which facilitates comparing and contrasting entities.
Typically, these models are designed to maximize the similarity 
between two different augmentations of the same image, 
while employing various regulations to avoid the collapse problem 
of converging to a trivial constant solution. 
Contrastive learning is an effective approach to avoid undesired trivial solutions. Techniques such as SimCLR, MoCo, PIRL leverage this approach by contrasting between positive and negative pairs, e.g., pulling positive pairs closer while pushing negative pairs farther apart \cite{chenSimpleFrameworkContrastive2020,heMomentumContrastUnsupervised2020,misraSelfSupervisedLearningPretextInvariant2019,wu2018unsupervised}.
Another approach SwAV \cite{caronUnsupervisedLearningVisual2020} utilizes online clustering 
to prevent trivial solutions, which clusters features to
prototypes while enforcing consistency between cluster assignments of different views of the same image.
Without using negative pairs for explicit contrasting,
asymmetric structure and momentum encoder have been considered for preventing collapsing \cite{grillBootstrapYourOwn2020,chenExploringSimpleSiamese2020,zhengReSSLRelationalSelfSupervised2021,fengAdaptiveSoftContrastive2022,heMomentumContrastUnsupervised2020,wangSelfSupervisedLearningEstimating2021}.
For example, BYOL uses a Siamese network with one branch being a momentum encoder
and directly predicts the output representation of one branch from another \cite{grillBootstrapYourOwn2020}. 
Then, it has been recognized in \cite{chenExploringSimpleSiamese2020} that the momentum encoder in BYOL is
unnecessary for preventing collapsing, rather a stop-gradient operation 
is crucial for avoiding collapsing.
Furthermore, the Barlow Twins method \cite{zbontarBarlowTwinsSelfSupervised2021} 
shows that, without using negative pairs 
and asymmetry structure,
the collapse problem can be naturally avoided by feature-wise contrastive learning.
It maximizes the similarity 
between the embeddings of distorted versions of the same sample, 
while minimizing the redundancy between the features of the embeddings. 
Moreover, VICReg \cite{bardesVICRegVarianceInvarianceCovarianceRegularization2022} 
explicitly avoids the collapse problem using a regularization term 
on feature variance and combining it with redundancy reduction and 
covariance regularization to form a variance-invariance-covariance regularization
formulation. 

\subsubsection{Mutual Information Maximization For Self-Supervised Learning}
MI is a fundamental quantity for measuring the dependence between random variables based on Shannon entropy. 
While calculating MI has traditionally been challenging, neural network-based methods have been recently developed to estimate MI for high-dimensional random variables \cite{belghazi2018mutual,gutmann2010noise}.
For SSL, many early methods \cite{wu2018unsupervised,oord2018representation,chenSimpleFrameworkContrastive2020,heMomentumContrastUnsupervised2020} use a contrastive loss function called InfoNCE or its variants. 
These methods commonly employ an NCE estimation of MI based on discriminating between positive and negative pairs \cite{gutmann2010noise}. These methods are linked to MI maximization as the NCE loss is a lower bound of MI \cite{poole2019variational}.
As the NCE estimator of MI is low-variance but high-bias,
a large batch size is required at test time for accurate 
MI estimation when the MI is large.
Moreover, in \cite{liuSelfSupervisedLearningMaximum2022}, 
the minimal coding length in lossy coding is used as a 
surrogate to construct a maximum entropy coding objective for SSL.
Additionally, the recent work \cite{shwartz2023information} has shown that
the VICReg method \cite{bardesVICRegVarianceInvarianceCovarianceRegularization2022} in fact implements an approximation of MI maximization criterion.
Besides, MI maximization is considered in \cite{ozsoy2022self} for
SSL using a log-determinant approximation of the MI,
which is then simplified to derive a Euclidean distance-based objective.
While these methods can be viewed as approximate implementations of the
MI maximization criterion, we consider explicit MI maximization for SSL.

%% file: SM/ProofTh1.tex
\subsection{Proof of Theorem 1} \label{app:A}

We first recall the result on the invariance property of mutual information. Specifically, if $Y = F(Z)$ and $Y' = G(Z')$ are homeomorphisms, then $I(Z;Z') = I(Y;Y')$ \cite{kraskov2004estimating}. Denote $\tilde Y = {[{Y^T},{Y'^T}]^T}$, if $Y = F(Z)$ and $Y' = G(Z')$ are Gaussian distributed, i.e. $Y \sim \mathcal{N}(Y;{\mu _Y},{C_{YY}})$ and $Y' \sim \mathcal{N}(Y';{\mu _{Y'}},{C_{Y'Y'}})$, we have $\tilde Y \sim \mathcal{N}(\tilde Y;{\mu _{\tilde Y}},{C_{\tilde Y\tilde Y}})$ with 
\[{C_{\tilde Y\tilde Y}} = \left[ {\begin{array}{*{20}{c}}
{{C_{YY}}}&{{C_{YY'}}}\\
{{C_{Y'Y}}}&{{C_{Y'Y'}}}
\end{array}} \right].\]
Then, the mutual information $I(Y;Y') = H(Y) + H(Y') - H(Y,Y')$ is given by
\[I(Y;Y') = \frac{1}{2}\log \frac{{\det ({C_{YY}})\det ({C_{Y'Y'}})}}{{\det ({C_{\tilde Y\tilde Y}})}},\]
where $H(Y)$ and $H(Y')$ are the marginal entropy of $Y$ and $Y'$, respectively, $H(Y,Y')$ is the joint entropy of   $Y$ and $Y'$. This together with the invariance property of mutual information, i.e. $I(Z;Z') = I(Y;Y')$ under homeomorphisms condition, results in Theorem 1.

\subsection{Proof of Theorem 2}\label{app:B}
\begin{figure}[b]
\vspace{-4mm}
    \centering
    \includegraphics[height=0.25\textwidth,width=0.35\textwidth]{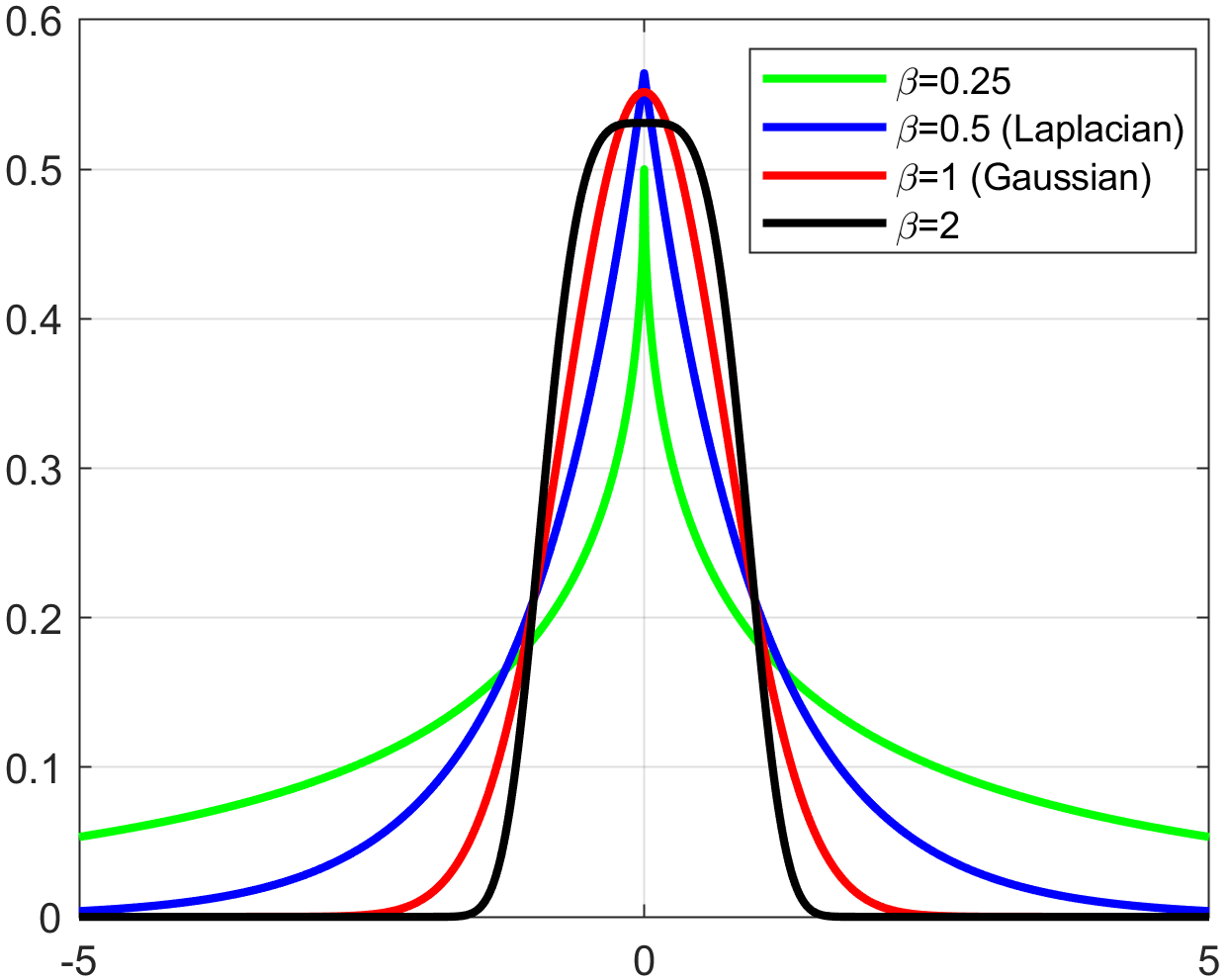}
    \caption{Univariate generalized Gaussian distribution with different values of the shape parameter. 
    }
    \label{fig:ggd}
    \vspace{-4mm}
\end{figure}

 Theorem 1 implies that we can compute the MI only based on second-order statistics even if the distributions of $Z$ and $Z^{\prime}$ are not Gaussian. We investigate the MI under the generalized Gaussian distribution (GGD) as defined in (3) of the main paper. The GGD offers a flexible parametric form that can adapt to a wide range of distributions by varying the shape parameter $\beta$ in (3), from super-Gaussian when $\beta<1$ to sub-Gaussian when $\beta>1$, including the Gamma, Laplacian and Gaussian distributions as special cases. Figure \ref{fig:ggd} provides an illustration of univariate GGD with different values.
%

\vspace{2mm}
Let $\tilde{Z}=\left[Z^{T}, Z^{\prime T}\right]^{T} \in \mathbb{R}^{2 d}$, and from (3) in main paper, the joint distribution ${p_{Z,Z'}}(z,z')$ is $\tilde{Z} \sim \mathcal{G N}\left(\tilde{Z} ; \mu_{\tilde{Z}}, \Sigma_{\tilde{Z} \tilde{Z}}, \beta\right)$, where $\mu_{\tilde{Z}}$ is the mean, and $\Sigma_{\tilde{Z} \tilde{Z}}$ is the dispersion matrix. The MI between $Z$ and $Z^{\prime}$ is given by
\begin{align*}
& I\left(Z, Z^{\prime}\right)\\
&=\iint p_{Z, Z^{\prime}}\left(z, z^{\prime}\right) \log \frac{p_{Z, Z^{\prime}}\left(z, z^{\prime}\right)}{p_Z(z) p_{Z^{\prime}}\left(z^{\prime}\right)} d z d z^{\prime} \tag{12} \\ 
& =E\left[\log p_{Z, Z^{\prime}}\left(z, z^{\prime}\right)\right]-E\left[\log p_Z(z)\right]-E\left[\log p_{Z^{\prime}}\left(z^{\prime}\right)\right].
\end{align*}
Then, it follows that
$$
\begin{aligned}
& E\left[\log p_{Z, Z^{\prime}}\left(z, z^{\prime}\right)\right] \\
& =\log \frac{\Phi(\beta, 2 n)}{\left[\operatorname{det}\left(\Sigma_{\tilde{Z}\tilde{Z}}\right)\right]^{1 / 2}}-\frac{1}{2} E\left\{\left[\left(\tilde{Z}-\mu_{\tilde{Z}}\right)^{T} \Sigma_{\tilde{Z} \tilde{Z}}^{-1}\left(\tilde{Z}-\mu_{\tilde{Z}}\right)\right]^{\beta}\right\}
\end{aligned},
$$
where the expectation over the parameter space $\mathbb{R}^{n}$ of a function $\varphi\left((\tilde{Z}-\mu_{\tilde{Z}})^{T} \Sigma_{\tilde{Z}\tilde{Z}}^{-1}\left(\tilde{Z}-\mu_{\tilde{Z}}\right)\right)=\varphi\left(\Bar{Z}^{T} \Bar{Z}\right) \equiv \varphi(w)$ (with $w>0$ for $\left.\tilde{Z}-\mu_{\tilde{Z}} \neq 0\right)$ is essentially type 1 Dirichlet integral, which can be converted into integral over $\mathbb{R}^{+}$ \cite{verdoolaegeGeometryMultivariateGeneralized2012}. Specifically, let $\varphi(w)=w^{\beta}$, then

$$
\begin{aligned}
& E\left\{\left[\left(\tilde{Z}-\mu_{\tilde{Z}}\right)^{T} \Sigma_{\tilde{Z} \tilde{Z}}^{-1}\left(\tilde{Z}-\mu_{\tilde{Z}}\right)\right]^{\beta} \right\}\\
& =\frac{\Phi(\beta, 2 n)}{\left[\operatorname{det}\left(\Sigma_{\tilde{Z} \tilde{Z}}\right)\right]^{1 / 2}} \int_{\mathbb{R}^{2 n}}\left[\left(\tilde{Z}-\mu_{\tilde{Z}}\right)^{T} \Sigma_{\tilde{Z} \tilde{Z}}^{-1}\left(\tilde{Z}-\mu_{\tilde{Z}}\right)\right]^{\beta}\\ &~~~~~~~\times\exp \left(-\frac{1}{2}\left[\left(\tilde{Z}-\mu_{\tilde{Z}}\right)^{T} \Sigma_{\tilde{Z} \tilde{Z}}^{-1}\left(\tilde{Z}-\mu_{\tilde{Z}}\right)\right]^{\beta}\right) d \tilde{z} \\
& \stackrel{(a)}{=} \frac{\beta}{2^{n / \beta} \Gamma(n / \beta)} \int_{\mathbb{R}^{+}} \varphi(w) w^{n-1} \exp \left(-\frac{1}{2} w^{\beta}\right) d w \\
& =\frac{\beta}{2^{n / \beta} \Gamma(n / \beta)} \frac{2^{2+n / \beta} \Gamma\left((\beta+n) / \beta\right)}{2 \beta} \\
& =\frac{2 \Gamma((\beta+n) / \beta)}{\Gamma(n / \beta)}\\
&=\frac{2 n}{\beta}
\end{aligned},
$$

where (a) is due to the fact that the density function of the positive variable $w=\left(\tilde{Z}-\mu_{\tilde{Z}}\right)^{T} \Sigma_{\tilde{Z} \tilde{Z}}^{-1}\left(\tilde{Z}-\mu_{\tilde{Z}}\right)$ is given by \cite{verdoolaegeGeometryMultivariateGeneralized2012}

$$
p(w ; \beta)=\frac{\beta}{\Gamma\left(\frac{n}{\beta}\right) 2^{\frac{n}{\beta}}} w^{n-1} \exp \left(-\frac{1}{2} w^{\beta}\right).
$$

Therefore,

$$
E\left[\log p_{Z, Z^{\prime}}\left(z,z ^{\prime}\right)\right]=\log \frac{\Phi(\beta, 2 n)}{\left[\operatorname{det}\left(\Sigma_{\tilde{Z} \tilde{Z}}\right)\right]^{1 / 2}}-\frac{2 n}{\beta}.
$$

Similarly,

$$
\begin{gathered}
E\left[\log p_{Z}(z)\right]=\log \frac{\Phi(\beta, n)}{\left[\operatorname{det}\left(\Sigma_{ZZ}\right)\right]^{1 / 2}}-\frac{n}{\beta}, \\
E\left[\log p_{Z^{\prime}}\left(z^{\prime}\right)\right]=\log \frac{\Phi(\beta, n)}{\left[\operatorname{det}\left(\Sigma_{Z^{\prime} Z^{\prime}}\right)\right]^{1 / 2}}-\frac{n}{\beta}.
\end{gathered}
$$

Substituting these expectations into (7) in main paper leads to

\begin{align*}
& I\left(Z, Z^{\prime}\right) \\
& =\frac{1}{2} \log \frac{\operatorname{det}\left(\Sigma_{Z Z}\right) \operatorname{det}\left(\Sigma_{Z^{\prime} Z^{\prime}}\right)}{\operatorname{det}\left(\Sigma_{\tilde{Z} \tilde{Z}}\right)}+\log \frac{\Phi(\beta, 2 n)}{[\Phi(\beta, n)]^{2}}  \tag{13}\\
& =\frac{1}{2} \log \frac{\operatorname{det}\left(\Sigma_{Z Z}\right) \operatorname{det}\left(\Sigma_{Z^{\prime} Z^{\prime}}\right)}{\operatorname{det}\left(\Sigma_{\tilde{Z} \tilde{Z}}\right)},
\end{align*}

where we used $\Phi(\beta, n)=\frac{\beta \Gamma(n / 2)}{2^{n /(2 \beta)} \pi^{n / 2} \Gamma(n /(2 \beta))}$ and the following relation

$$
\frac{\Phi(\beta, 2 n)}{[\Phi(\beta, n)]^{2}}=\frac{\beta \Gamma(n)}{\Gamma(n / \beta)} \frac{\left[\Gamma\left(\frac{n}{2 \beta}\right)\right]^{2}}{\left[\beta \Gamma\left(\frac{n}{2}\right)\right]^{2}}=\frac{1}{\beta} \frac{2^{n-\frac{1}{2}}}{2^{\frac{n}{\beta}-\frac{1}{2}}} \frac{\Gamma\left(\frac{n}{2}+\frac{1}{2}\right)}{\Gamma\left(\frac{n}{2 \beta}+\frac{1}{2}\right)} \frac{\Gamma\left(\frac{n}{2 \beta}\right)}{\Gamma\left(\frac{n}{2}\right)}=1.
$$

Then, using the relation between the dispersion matrix and covariance matrix

\begin{equation*}
{\Sigma _{\tilde X\tilde X}}{\rm{ = }}\frac{{n\Gamma (n/(2\beta ))}}{{{2^{1/\beta }}\Gamma ((n + 2)/(2\beta ))}}{C_{\tilde X\tilde X}},
\end{equation*}

it follows that

$$
I\left(Z ; Z^{\prime}\right)=\frac{1}{2} \log \frac{\operatorname{det}\left(\Sigma_{ZZ}\right) \operatorname{det}\left(\Sigma_{Z^{\prime} Z^{\prime}}\right)}{\operatorname{det}\left(\Sigma_{\tilde{Z} \tilde{Z}}\right)}=\frac{1}{2} \log \frac{\operatorname{det}\left(C_{Z Z}\right) \operatorname{det}\left(C_{Z^{\prime} Z^{\prime}}\right)}{\operatorname{det}\left(C_{\tilde{Z} \tilde{Z}}\right)}.
$$

%% file: SM/Ablations.tex
\subsection{Ablation Study}  \label{app:Ablation}
\subsubsection{Loss Function}
We investigate the effectiveness of each term of the loss function. Specifically, we remove one of $\log{\operatorname{det}{C_{ZZ}}}$ term (w/o $\log{\operatorname{det}{C_{ZZ}}}$) and the $\log{\operatorname{det}{C_{Z'Z'}}}$ term (w/o $\log{\operatorname{det}{C_{Z'Z'}}}$) or both terms (w/o  both) from the loss function. Additionally, we replace the $\log{\operatorname{det}({C_{ZZ}-C_{Z'Z'}})}$ term with $\|Z-Z'\|^2$ since both terms aim to align the representations $Z$ and $Z'$. Furthermore, we simplify the rescaling operation from $ \tilde{M}=\frac{M-\mu_{\lambda} I}{\alpha}+I$ to $ \tilde{M}=\frac{M}{\alpha}+I$ (w/o $\mu_{\lambda}$).
Results in Table \ref{tab:abla_variants} show that removing either $\log{\operatorname{det}{C_{ZZ}}}$ or $\log{\operatorname{det}{C_{Z'Z'}}}$ leads to performance decrease, yet the training still succeed. However, removing both leads to training failure. 
\begin{table} 
  \caption{\textbf{Ablation on loss function. } The experiment follows the same setup as in Table \ref{tab:CIFAR_linear}
 in main paper, with Top-1 accuracy is reported. 
  }
  \label{tab:abla_variants}
  \centering
  \begin{tabular}{lcccc}
    \toprule
    Method & CIFAR-100  & ImageNet-100 \\
    \midrule
    Original & 70.5 &  81.1 \\
     w/o  $\log{\operatorname{det}{C_{ZZ}}} $ &  66.6 &  76.5 \\
     w/o  $\log{\operatorname{det}{C_{Z'Z'}}}$ & 67.7 &  78.8 \\
    w/o both & 3.55 &  4.01 \\
    Using $\|Z-Z'\|^2$ & 69.8 &  79.6 \\
         w/o  $\mu_{\lambda}$  in rescaling   &  70.6 &  80.3 \\ 
    \bottomrule
  \end{tabular}
\end{table}
The reason behind this is straightforward. By minimizing the term $\log{\operatorname{det}({C_{ZZ}-C_{Z'Z'}})}$, we aim to align the representations $Z$ and $Z'$. The terms $\log{\operatorname{det}{C_{ZZ}}}$ and $\log{\operatorname{det}{C_{Z'Z'}}}$ play a crucial role in ensuring that these representations are informative enough to prevent representation collapse. 
When one of these terms is removed, the remaining term is expected to partially fulfill this, but becomes less effective.

It can be seen from Table \ref{tab:abla_variants} that replacing the $\log{\operatorname{det}({C_{ZZ}-C_{Z'Z'}})}$ term with $\|Z-Z'\|^2$ decreases the performance.
Although both $\log{\operatorname{det}({C_{ZZ}-C_{Z'Z'}})}$ and  $\|Z-Z'\|^2$ encourage consistency between $Z$ and $Z'$, their mathematical properties differ significantly. The term $\log{\operatorname{det}({C_{ZZ}-C_{Z'Z'}})}$  encourages a holistic consistency in the structural properties of the feature spaces. Meanwhile, our derived loss function obviates the need to tune the balance ratio between the terms. Moreover, the results show that removing $\mu_{\lambda}$ term in the Taylor approximation, i.e. adding a fixed $I_d$ to the three terms in the loss function does not affect the performance on CIFAR-100 but decreases the performance on ImageNet-100. 

\subsubsection{Projector Hidden Dimension and Projector Output Dimension}
We evaluate the effect of the projector's hidden dimension and projector output dimension in Table \ref{tab:abl_pj_hidden}. For both our method and Barlow Twins, there's a tendency that the increase of projector hidden dimension generally improves the performance on both the CIFAR-100 and ImageNet-100 datasets. 
Compared with CIFAR-100, both our method and Barlow Twins need a larger 
hidden dimension to achieve high performance on the more complex ImageNet-100 dataset. Using a momentum encoder, our method can achieve better performance
and becomes more robust to projector hidden dimension.
Moreover, similar to the results on the hidden dimension, 
both our method and Barlow Twins exhibit a trend that 
increasing the projector output dimension generally improves performance. 
The performance of Barlow Twins is particularly sensitive to projector output dimension, whereas our method performs well even with a very small projector output dimension of 256.  

\begin{table}[h]
  \caption{\textbf{Impact of projector hidden/output dimension on accuracy.}}
  \label{tab:abl_pj_hidden}
  \centering
  \begin{tabular}{ccccccc}
    \toprule
  & \multicolumn{3}{c}{{CIFAR100}} & \multicolumn{3}{c}{{ImageNet100}} \\
  \cmidrule(r){1-4} \cmidrule(r){5-7}
        {Proj. hidden dim} & {Ours} & {Ours-M} & {Barlow Twins} & {Ours} & {Ours-M} & {Barlow Twins} \\
    \midrule
    2048 & 70.5 & 70.4 & 70.9 & 81.1 & 81.7 & 80.4 \\
1024 & 70.8 & 70.4 & 70.2 & 80.2 & 81.5 & 79.3 \\
512  & 69.1 & 70.1 & 69.6 & 79.5 & 81.4 & 78.3 \\
256  & 67.9 & 70.0 & 68.0 & 78.7 & 80.6 & 76.9 \\
     \cmidrule(r){1-4} \cmidrule(r){5-7}
    {Proj. output dim} & {Ours} & {Ours-M} & {Barlow Twins} & {Ours} & {Ours-M} & {Barlow Twins}    \\
    \midrule
    2048 & 70.5 & 70.4 & 70.9 & 81.1 & 81.7 & 80.4 \\
1024 & 70.3 & 70.4 & 69.7 & 80.6 & 81.1 & 79.6 \\
512  & 70.6 & 70.5 & 66.5 & 80.4 & 81.7 & 77.4 \\
256  & 70.5 & 71.1 & 62.1 & 80.3 & 81.2 & 73.6 \\
\bottomrule
  \end{tabular}
\end{table}

%% file: SM/exp_details.tex
\subsection{ Experiment Implementation Details} \label{app:C}
For experiments on ImageNet-1K, we use a batch size of 1020 on 3 A100 GPUs for 100, 400, and 800 epochs. 
Training is conducted using 16-bit precision (FP16) and 4 batches of gradient accumulation to stabilize model updating and accelerate the training process. We use the LARS optimizer with a base learning rate of 0.8 for the backbone pretraining and 0.2 for the classifier training. The learning rate is scaled by $ \text{lr} = \text{base\_lr}  \times \text{batch\_size}  /  256 \times \text{num\_gpu} $.  We use a weight decay of 1.5E-6 for backbone parameters. 
The linear classifier is trained on top of frozen backbone. We follow the default setting as in Solo-learn benchmark \cite{costaSololearnLibrarySelfsupervised2022} for the rest of the training hyper-parameters.  

\begin{figure}  
\vspace{-0mm}
    \centering
    \includegraphics[
    width=0.35\textwidth]{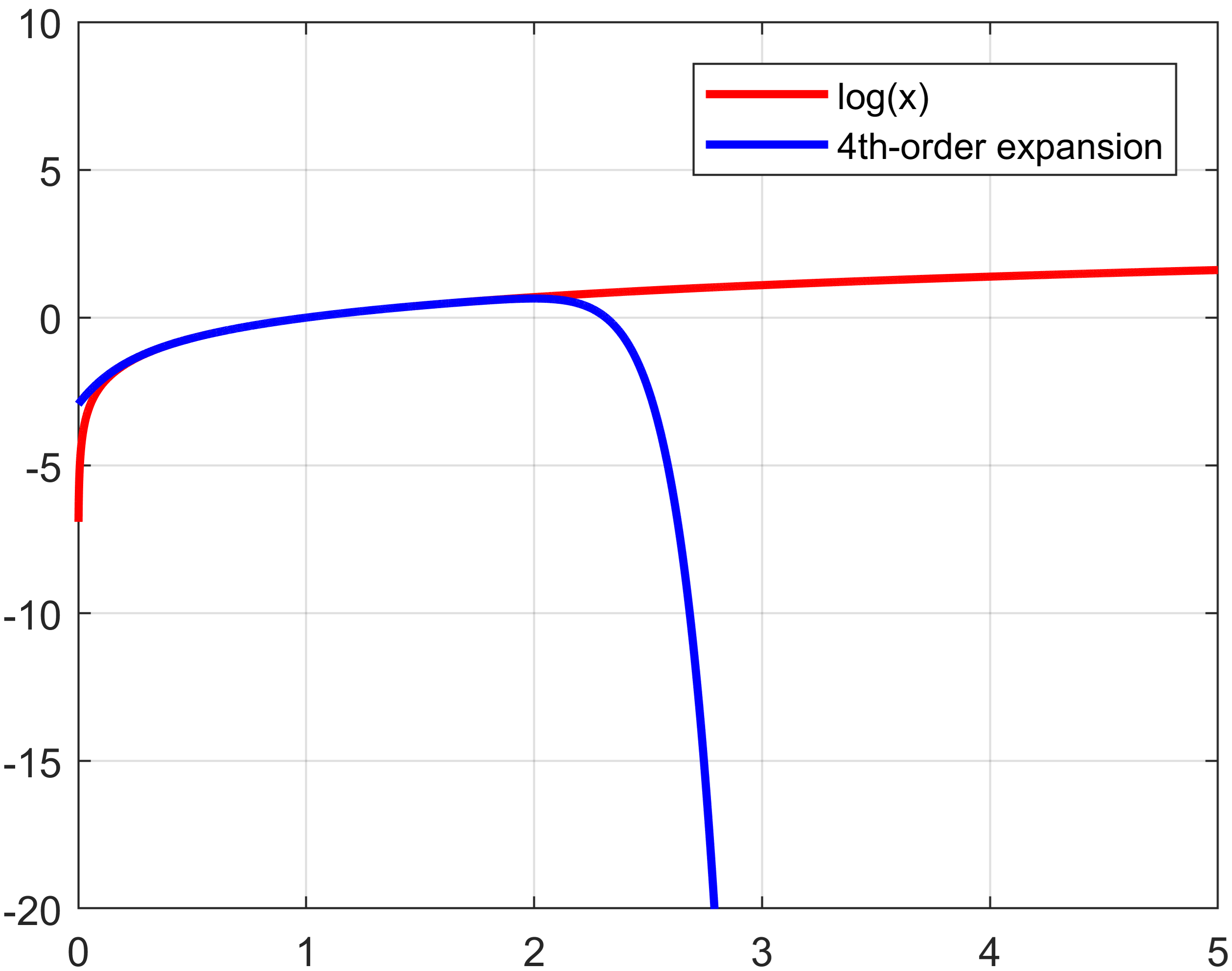}
    \caption{Illustration of a fourth-order approximation of the log function in (10) in main paper.
    }
    \label{fig:tylor}
\vspace{-4mm}
\end{figure}

 Recall that in implementing the loss function (9) from the main paper, the three log-determinant terms are expanded as 
 \begin{align*}\label{tyler_expan}
 \log \operatorname{det}(M)&=\sum_{i=1}^{n} \log \lambda_{i}(M) \\
& =\sum_{i=1}^{n} \sum_{k=1}^{\infty}(-1)^{k+1} \frac{\left(\lambda_{i}(M)-1\right)^{k}}{k} \\
& =\sum_{k=1}^{\infty}(-1)^{k+1} \frac{\operatorname{tr}\left((M-I)^{k}\right)}{k}=\operatorname{tr}\left(\sum_{k=1}^{\infty}(-1)^{k+1} \frac{(M-I)^{k}}{k}\right),\tag{10}
\end{align*}retaining only a $p$-th order approximation is kept, e.g., $p=4$ in the experiments of this work. As shown in Figure \ref{fig:tylor}, a fourth-order approximation of the log function in (10) is sufficiently accurate around the value of 1.

\begin{figure} 
    \centering
    \includegraphics[width=0.4\linewidth]{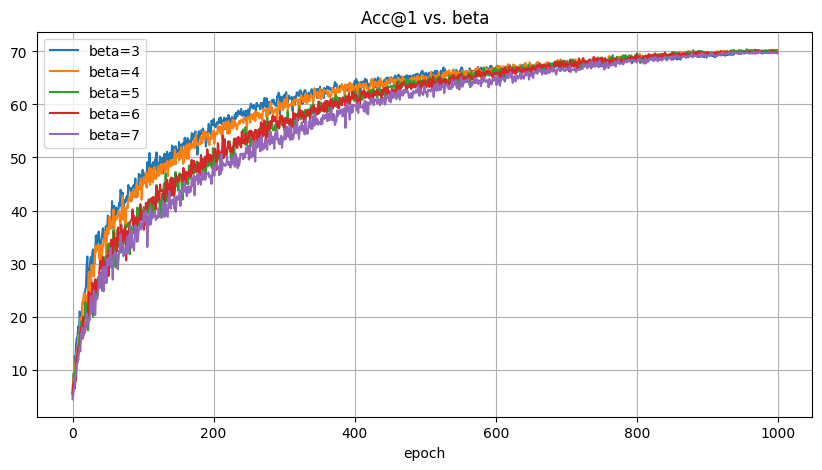}
    \caption{The convergence curves of our method on CIFAR-100 for different values of the parameter $\beta$ used for the rescaling operation.}
    \label{fig:abl_beta}
\end{figure}

For our method with a momentum encoder, we follow the setting of \cite{grillBootstrapYourOwn2020,chenExploringSimpleSiamese2020} 
and use a two-layer predictor with hidden dimension 1024 for all datasets. 
For ImageNet-100, we set the base learning rate as 0.2 for backbone pretraining and 0.3 for the classifier. We set the weight decay of backbone parameters as 0.0001. 
For CIFAR-100, we set the base learning rate for backbone pretraining as 0.3 and the classifier as 0.2. The weight decay is set as 6E-5. For the rescaling operation ($\tilde{M}=\frac{M-\mu_{\lambda} I}{\alpha}+I $), we track the eigenvalues of $C_{ZZ}$ with an update interval of 100 batches with a moving average coefficient $\rho$ of 0.99.
Table \ref{tab:update-ema-CIFAR-100-Table} shows the ablation study on the update interval and moving average coefficient $\rho$. Generally, a larger update interval should be used with a smaller moving average coefficient, and vice versa. This is reasonable as the two parameters together control the speed of the eigenvalue tracking.
Overall, our method is insensitive to these two hyperparameters.  
Table \ref{tab:CIFAR_linear}
 in main paper and Figure \ref{fig:abl_beta} depict the ablation study on the parameter $\beta$ used for the rescaling operation. We adhere to the experimental setup described in Table \ref{tab:CIFAR_linear}
 and report the Top-1 accuracy on CIFAR-100.  As shown in Table \ref{tab:beta-cifar100} and Figure \ref{fig:abl_beta}, a smaller $\beta$ results in faster convergence during training. However, excessively small values may lead to training failure. We set $\beta=5$ for all the experiments.
\begin{table}
    \caption{Ablation study on the update interval and moving average coefficient $\rho$ for eigenvalues tracking used for the rescaling operation.}
    \label{tab:update-ema-CIFAR-100-Table}
    \centering
    \begin{tabular}{lcc}
      \toprule
      update interval & $\rho$ & CIFAR-100 \\
      \midrule
      1000 & 0 & 69.45 \\
      1000 & 0.1 & 70.22 \\
      1000 & 0.99 & 68.69 \\
      100 & 0.99 & 70.54 \\
      1 & 0.99 & 69.17 \\
      1 & 0 & 69.37 \\
      \bottomrule
    \end{tabular}
  \end{table}
  
\begin{table} 
  \caption{{Ablation study on the parameter $\beta$ used for the rescaling operation }. }
  \label{tab:beta-cifar100}
  \centering
  \begin{tabular}{cc}
    \toprule
   ${\beta}$ & {CIFAR-100} \\
    \midrule
    7  & 69.85 \\
    6  & 70.10 \\
    5  & 70.17 \\
    4  & 70.09 \\
    3  & 69.46 \\
    1  &$ \mathrm{NaN} $\\
    \bottomrule
  \end{tabular}
\end{table}

\clearpage